\documentclass[]{acmart}

\AtBeginDocument{%
  \providecommand\BibTeX{{%
    \normalfont B\kern-0.5em{\scshape i\kern-0.25em b}\kern-0.8em\TeX}}}

\setcopyright{acmcopyright}
\copyrightyear{2022}
\acmYear{2022}
\acmDOI{XXX}

\copyrightyear{2022}
\acmYear{2022}
\setcopyright{rightsretained}
\acmConference[FAccT '22]{2022 ACM Conference on Fairness, Accountability, and Transparency}{June 21--24, 2022}{Seoul, Republic of Korea}
\acmBooktitle{2022 ACM Conference on Fairness, Accountability, and Transparency (FAccT '22), June 21--24, 2022, Seoul, Republic of Korea}\acmDOI{10.1145/3531146.3533122}
\acmISBN{978-1-4503-9352-2/22/06}

\usepackage{url}
\usepackage{pdfpages}
\usepackage{wrapfig}
\usepackage{todonotes}
\usepackage{color}

\begin{document}

\title{REAL ML: Recognizing, Exploring, and Articulating Limitations of Machine Learning Research}


\author{Jessie J. Smith}
\affiliation{
\institution{University of Colorado Boulder}
\city{Boulder, CO}
\country{USA}
}
\email{jessie.smith-1@colorado.edu}

\author{Saleema Amershi}
\affiliation{
\institution{Microsoft Research}
\city{Redmond, WA}
\country{USA}
}
\email{samershi@microsoft.com}

\author{Solon Barocas}
\affiliation{
\institution{Microsoft Research}
\city{New York, NY}
\country{USA}
}
\email{solon.barocas@microsoft.com}

\author{Hanna Wallach}
\affiliation{
\institution{Microsoft Research}
\city{New York, NY}
\country{USA}
}
\email{wallach@microsoft.com}

\author{Jennifer Wortman Vaughan}
\affiliation{
\institution{Microsoft Research}
\city{New York, NY}
\country{USA}
}
\email{jenn@microsoft.com}

\renewcommand{\shortauthors}{Smith et al.}

\begin{abstract}
  Transparency around limitations can improve the scientific rigor of research, help ensure appropriate interpretation of research findings, and make research claims more credible.  Despite these benefits, the machine learning (ML) research community lacks well-developed norms around disclosing and discussing limitations.  To address this gap, we conduct an iterative design process with 30 ML and ML-adjacent researchers to develop and test REAL ML, a set of guided activities to help ML researchers recognize, explore, and articulate the limitations of their research.  Using a three-stage interview and survey study, we identify ML researchers' perceptions of limitations, as well as the challenges they face when recognizing, exploring, and articulating limitations.
We develop REAL ML to address some of these practical challenges, and highlight additional cultural challenges that will require broader shifts in community norms to address.
We hope our study and REAL ML help move the ML research community toward more active and appropriate engagement with limitations.\looseness=-1
\end{abstract}




\maketitle

\section{Introduction}
\label{sec:intro}

Machine learning (ML) has emerged as one of the most active and impactful fields of research within computer science. The past decade has been marked by a steady stream of impressive technical achievements from the ML research community and many real-world applications of ML. These developments have fostered considerable hype about ML's potential, as well as growing concerns about the inflated and unsupported claims that are sometimes made about its true capabilities \cite{narayanan2019recognize}. In this paper, we consider the role that ML researchers can play in ensuring appropriate interpretation of their research. In particular, we focus on the disclosure and discussion of limitations---and why such practices are important to the healthy functioning of a research community and the broader understanding of its achievements.\looseness=-1

Limitations are drawbacks in the design or execution of research that may impact the resulting findings and claims. Limitations are different from ``research ethics,'' which focuses on the potential harms that may be caused to human subjects during the research process \cite{sim2021thinking}, and ``broader impacts,'' which focus on the potential downstream harms and consequences for society that may arise as a result of research findings and claims~\cite{nanayakkara2021unpacking,PAI-risks-of-research}. In contrast, limitations concern aspects of the research process that may pose a threat to the validity of research findings and claims.

Many scientific fields have well-established norms around disclosing and discussing limitations, which are often recognized as necessary for improving scientific rigor and research integrity. These norms rest on a shared belief that recognizing, exploring, and articulating limitations can foster greater precision in the descriptions of research (making it easier to reproduce), help ensure appropriate interpretation of research findings, make research claims more credible, and highlight issues that would benefit from further research \cite{christensen2018transparency, makel2017toward, burlig2018improving}. Although practices necessarily vary by field and by publication venue \cite{ioannidis2007limitations}, the ML research community is notable for not having particularly well-developed norms around disclosing and discussing limitations. Although there have been recent efforts to encourage ML researchers to reflect on the potential impacts (intended or not) of their research on society---for example, the introduction of broader impacts statements at the Neural Information Processing Systems (NeurIPS) conference in 2020 \cite{neurips-broaderimpact} and the subsequent NeurIPS 2021 paper checklist, which encouraged authors to articulate the limitations of their research and even to create separate limitations sections in their papers \cite{neurips-paperchecklist}---disclosure and discussion of limitations remains uncommon in the ML research community. \looseness=-1

In this paper, we present the Recognizing, Exploring, and Articulating
Limitations in Machine Learning tool (REAL ML), a set of guided
activities to help ML researchers recognize, explore, and articulate
the limitations of their research. We developed and tested REAL ML via
an iterative design process with 30 ML and ML-adjacent
researchers. Specifically, using a three-stage interview and survey
study, we 1) identified ML researchers' perceptions of limitations, as
well as the practical and cultural challenges they face when
recognizing, exploring, and articulating limitations; and 2)
iteratively developed REAL ML to address the some of the practical
challenges we identified. Additionally, we introduce a list of sources
of limitations and a list of types of limitations that commonly occur
in ML research, both of which were curated and refined over the course
of our study. Our findings reveal many challenges faced by ML
researchers, and show early evidence suggesting that REAL ML may help
them feel more prepared and more willing to recognize, explore, and articulate the limitations of their research. Our study also exposes cultural
challenges that go beyond the scope of REAL ML and will require
broader shifts in community norms to address. We hope our study and
REAL ML help move the ML research community toward more actively and
appropriately engaging with limitations.\looseness=-1

\section{Background and Related Work}
\label{sec:relatedwork}

Disclosing and discussing limitations can benefit both those doing research and those building on the research of others. Since limitations affect the validity of research findings and claims, there can be negative consequences when they are not recognized, explored, and articulated. Due to the severity of the potential harms that may arise if research findings are misinterpreted or taken out of context and the need to guard against both unintentional misreporting and deliberate spinning of research claims~\cite{boutron2018misrepresentation}, some publication venues in fields like biomedicine require limitations to be stated in papers' abstracts~\cite{YRH+14,G06}.  The Journal of the American Medical Association advises authors to include \emph{``a discussion section placing the results in context... and addressing study limitations''} \cite{JAMA}, while the Journal of Neuroscience requires authors to include \emph{``a discussion of the validity of the observations''} \cite{JNeuroSci}. However, these practices are still relatively rare; one recent study reported that only one of the 25 top-cited scientific journals encourages disclosure and discussion of limitations \cite{ioannidis2007limitations}, and, to the best of our knowledge, limitations sections are not yet required by any major ML publication venue.\looseness=-1

Even when limitations are disclosed, researchers often fail to appropriately discuss how these limitations might impact their research findings and claims~\cite{puhan2012discussing,ross2019limited}. \citet{lingard2021art} described three common approaches that researchers take to writing limitations sections: the confessional, where researchers beg forgiveness for their work's flaws from reviewers; the dismissal, where researchers acknowledge limitations only to diminish and dismiss their importance; and the reflection, where researchers acknowledge the uncertainty and assumptions that underlie their research, and reflect on their impacts.  Although the latter approach is ideal, the first two are common in practice.  There are a variety of reasons for this.  Perhaps chief among them is the perceived stigma around disclosing limitations and the fear that doing so would increase the likelihood of paper rejection. \citet{brutus2013self} observed that in the field of management, \emph{``the pressure stemming from the increasingly low acceptance rates for peer reviewed journals and the emphasis on publications in academic reward structures represent clear motives for not acknowledging limitations and for offering only benign directions for future research.''} Similarly, \citet{puhan2012discussing} noted that in biomedicine, researchers are reluctant to disclose limitations because \emph{``they perceive a transparency threshold beyond which the probability of manuscript acceptance goes down (perhaps even to zero).''} In our study, we see evidence of a similar perceived stigma within the ML research community.\looseness=-1

Another reason why researchers may fail to appropriately disclose or discuss limitations is a lack of guidance and appropriate training \cite{ioannidis2007limitations}. Although there are no widely agreed-upon approaches for recognizing, exploring, and articulating limitations, guidelines have begun to emerge in some fields. When developing REAL ML, we drew on a four-step process for articulating limitations in medical studies, proposed by  \citet{ross2019limited}.  This process involves stating whether each limitation arises in the study design, data collection, data analysis, or study results; explaining the potential impacts of the limitation; providing potential alternative approaches that could have been taken and why they were not; and describing any steps that were taken to mitigate the limitation's impacts. Within ML, researchers have begun to create tools and resources to help both researchers and practitioners think more critically about the impacts of their work, such as value cards \cite{shen2021value}, broader impacts statements \cite{neurips-broaderimpact}, model cards \cite{mitchell2019model}, and datasheets for datasets \cite{gebru2021datasheets}. REAL ML similarly encourages reflection, but we place less emphasis on ethical impacts and instead focus on the impacts of limitations on research findings and claims, looking across all aspects of ML research rather than just models and datasets. Importantly, REAL ML is intended for use \emph{posthoc} via a process known as ``reflection on action'' \cite{munby1989reflection}, a contemplative practice that uses reflection on previous actions to gain a better understanding of their impacts. Given the cultural challenges faced by ML researchers, we believe this is a necessary step in moving the ML research community toward ``reflection \emph{in} action''  \cite{schon1984architectural}, where informed trade-offs and just-in-time adjustments are made during the research process to reduce the likelihood of negative impacts and improve scientific rigor and research integrity; we return to this topic in Section~\ref{sec:discussion}. \looseness=-1

Although encouraging reflection is an important step toward normalizing disclosure and discussion of limitations, additional obstacles remain. Community norms
may hinder ML researchers' abilities to recognize, explore, and articulate the limitations of their research. As discussed by \citet{giraud2011importance}, ML publication venues tend to favor positive results over negative results, with research failures rarely discussed openly. As noted by \citet{liao2021we}, who created a taxonomy of threats to internal and external validity from 107 papers surveying ML research, the use of benchmarks to assess progress in ML can place too much emphasis on outcomes at the expense of scientific inquiry. Finally, the ML review process---like those found in much of computer science---tends to emphasize abstraction and generalizability. Indeed, in an analysis of highly cited papers published at two major ML conferences, \citet{birhane2021values} found that the third most common value expressed in ML papers was generalizability. This emphasis may encourage ML researchers to stretch their research claims in inappropriate ways, rather than explicitly describing their limited scope \cite{Sb+19, d2020data}. Despite these community norms, disclosure and discussion of limitations is just as important in ML as in other scientific fields, particularly as ML research is often motivated by or meant to influence real-world applications \cite{wagstaff2012machine}.  \looseness=-1

Our findings provide insight into ML researchers' perceptions of limitations, as well as the practical and cultural challenges they face when recognizing, exploring, and articulating limitations. Although we highlight some challenges that stem from community norms, our primary goal is to provide practical support for ML researchers. REAL ML therefore contains structure to help ML researchers reflect on the ways that limitations arise from unavoidable constraints, unforeseen challenges, and decisions made during the ML research process, and on the impacts these limitations may have on the resulting findings and claims, with the aim of using this reflection to improve scientific rigor and research integrity.\looseness=-1

\section{Methods}
\label{sec:methods}

We conducted an iterative design process aimed at developing and testing REAL ML, a set of guided activities to help ML researchers recognize, explore, and articulate the limitations of their research. Specifically, we used a three-stage interview and survey study to answer the following four research questions:
\begin{itemize}
    \item \textbf{RQ1:} What is a limitation of ML research? What types of limitations are there? How do they arise?
    \item \textbf{RQ2:} What challenges do ML researchers face when seeking to recognize limitations?
    \item \textbf{RQ3:} What challenges do ML researchers face when exploring and articulating limitations?
    \item \textbf{RQ4:} What practical support would help alleviate some of these challenges for ML researchers?
\end{itemize}
As we explain in detail below, our study included interviews with 20 ML researchers while using evolving versions of REAL ML (stage 1), interviews and reviews of REAL ML with six ML-adjacent researchers (stage 2), and a final stage with four additional ML researchers who provided feedback via an online survey using a near-final version of REAL ML (stage 3). Throughout the study, we iterated on REAL ML based on the feedback we received from participants. All interviews were conducted virtually on a video conferencing platform.  Interviews were recorded and transcribed using third-party software. Participation was voluntary; each interview participant was compensated with a \$30 voucher, while each survey participant was compensated with a \$45 voucher. The study was approved by our institution's IRB.\looseness=-1

\paragraph{Initial prototype.}

Prior to beginning our study, we developed an initial prototype of REAL ML. When developing this prototype, we drew on our team's interdisciplinary expertise in ML, HCI, and science and technology studies, as well as our decades of combined experience writing and reviewing ML papers and engaging with the ML research community. The prototype consisted of three lists intended to encourage reflection: a list of types of limitations that commonly occur in ML research (e.g., generalizability limitations, robustness limitations) and descriptions of each; a list of common decision-making points in the ML research process where limitations could arise (e.g., formalism of the problem, technical approach) and descriptions of each; and a list of probing questions to answer when preparing to articulate a limitation (e.g., questions about potential alternative approaches that could have been taken, questions about how the limitation's impacts were mitigated).  The list of commonly occurring types of limitations drew on the limitations uncovered in Nanayakkara et al.'s analysis of NeurIPS 2020 broader impacts statements \cite{nanayakkara2021unpacking} and on Birhane et al.'s analysis of the values expressed in ML papers \cite{birhane2021values}. The probing questions were adapted from Ross and Zaidi's four-step process for articulating limitations in medical studies \cite{ross2019limited}. After constructing an initial version of each list, we piloted the lists using our own ML papers and updated them based on our experiences.  We additionally piloted the initial prototype with and solicited informal feedback from colleagues within and outside the ML research community and further iterated on the prototype based on this feedback. The initial prototype is included in the appendix. \looseness=-1

\paragraph{Stage 1.}
After creating our initial prototype, we conducted semi-structured interviews with 20 ML researchers. We recruited participants through social media, posting links to a recruitment form on our Twitter accounts. We specifically sought to recruit ML researchers who had previously published at least one peer-reviewed ML paper.  Although we obtained a relatively diverse sample of 100 researchers, we acknowledge that social media recruitment can lead to selection bias, so the researchers who expressed interest in participating in our study may have already been more likely to use a tool like REAL ML. Starting with the 100 researchers who completed our recruitment form, we filtered out researchers who had not both authored and reviewed at least one peer-reviewed ML paper (saving some for stage 2), binned the remaining researchers based on their years of experience with ML research and their geographic locations, and then randomly selected a fixed number of researchers from each bin, yielding 20 participants in total. We intentionally selected participants with different levels of experience and different geographic locations because we wanted to ensure REAL ML would be suitable for different research contexts. Table \ref{table:total-participants} contains more information about the participants. \looseness=-1

\begin{table*}
  \caption{Information about participants.}
  \label{table:total-participants}
\begin{tabular}{p{0.25cm}p{1cm}p{1.35cm}p{0.7cm}p{5.9cm}p{3.5cm}}
\toprule
    \textbf{ID} &
    \textbf{Country}&
    \textbf{Experience} &
    \textbf{Stage} &
    \textbf{Area(s) of expertise} &
    \textbf{Version of REAL ML}\\
\midrule
            P1 &     USA &  < 5 years & 1 &                  NLP, computational social science &     V1 Prototype with guidance \\
            \hline
            P2 &     USA &  < 5 years & 1 &          model based relational learning, robotics &     V2 Prototype with guidance \\
            \hline
            P3 &     USA & > 10 years & 1 &    NLP, information retrieval, medical informatics &     V2 Prototype with guidance \\
            \hline
            P4 &     USA & > 10 years & 1 & NLP, interpretable ML, reinforcement learning, AI ethics &     V2 Prototype with guidance \\
            \hline
            P5 &     USA & 5-10 years & 1 &                               NLP, computer vision &     V2 Prototype with guidance \\
            \hline
            P6 &   China &  < 5 years & 1 &                    medical informatics, healthcare &     V2 Prototype with guidance \\
            \hline
            P7 &   India &  < 5 years & 1 &                                        data mining &     V2 Prototype with guidance \\
            \hline
            P8 &   Spain &  < 5 years & 1 &         information retrieval, recommender systems &     V2 Prototype with guidance \\
            \hline
            P9 &   India &  < 5 years & 1 &                                 NLP, deep learning &     V2 Prototype with guidance \\
            \hline
           P10 &     USA & 5-10 years & 1 &                   AI fairness, recommender systems &     V3 Prototype with guidance \\
           \hline
           P11 & Germany & 5-10 years & 1 &                adversarial attacks, trustworthy AI &     V3 Prototype with guidance \\
           \hline
           P12 &     USA & 5-10 years & 1 &                                                NLP &     V3 Prototype with guidance \\
           \hline
           P13 &     USA & 5-10 years & 1 &                        NLP, reinforcement learning &          V1 Tool with guidance \\
           \hline
           P14 &     USA &  < 5 years & 1 &                    NLP, AI fairness, privacy in AI &          V1 Tool with guidance \\
           \hline
           P15 &  Canada & 5-10 years & 1 & explainability, human-centered AI, medical informatics &          V2 Tool with guidance \\
           \hline
           P16 &     USA & 5-10 years & 1 &                         NLP, structured prediction &          V2 Tool with guidance \\
           \hline
           P17 &     USA &  < 5 years & 1 &                                                NLP &          V2 Tool with guidance \\
           \hline
           P18 &     USA & 5-10 years & 1 &                                AI ethics, fairness &            V2 Tool no guidance \\
           \hline
           P19 &     USA & > 10 years & 2 &                                social science, HCI &                        V2 Tool \\
           \hline
           P20 &     USA & > 10 years & 2 & cognitive psychology, behavioral economics, actuarial decision making &                        V2 Tool \\
           \hline
           P21 &     USA & > 10 years & 2 & responsible AI, standardization, research ethics &                        V3 Tool \\
           \hline
           P22 &     USA & 5-10 years & 1 &                 learning theory, foundations of ML &            V3 Tool no guidance \\
           \hline
           P23 &     USA & > 10 years & 2 &                              HCI, social computing &                        V3 Tool \\
           \hline
           P24 &  Canada &  < 5 years & 1 &                                                NLP &            V3 Tool no guidance \\
           \hline
           P25 &     USA & > 10 years & 2 &            social science methods, ethics in AI/ML &                        V3 Tool \\
           \hline
           P26 &     USA & > 10 years & 2 &                     ethics in AI/ML, AI governance &                        V3 Tool \\
           \hline
           S1 &     USA & 5-10 years & 3 &                                                 -- &            V4 Tool no guidance \\
           \hline
           S2 &     USA &  < 5 years & 3 &                                                 -- &            V4 Tool no guidance \\
           \hline
           S3 &  Canada & > 10 years & 3 &                                                 -- &            V4 Tool no guidance \\
           \hline
           S4 &     USA & 5-10 years & 3 &                                                 -- &            V4 Tool no guidance \\
\bottomrule
\end{tabular}
\end{table*}

Prior to each participant's interview, we asked them to share with us a publicly available (e.g., published or available on arXiv) ML paper they had authored to use as a case study, with the guidance that the paper should fall into ``the topic areas covered by ML venues such as NeurIPS, ICML, ICLR, COLT, and AISTATS or related venues like ACL, EMNLP, or CVPR.'' The interviewer read each participant's paper before conducting their interview in order to provide personalized guidance on using REAL ML. We began each interview by asking the participant to reflect on their previous experiences recognizing, exploring, and articulating the limitations of their research, including any challenges they faced, and to provide their own definition of limitations of ML research. Next, we asked them to walk through the process of recognizing limitations in the paper they had shared with us, as if it had not yet been published, using the latest version of REAL ML. Several participants' papers already mentioned limitations, so we encouraged these participants to focus on new limitations they had not previously recognized.  With early versions of REAL ML (labeled as ``Prototype'' in Table~\ref{table:total-participants}, last column) that lacked any guiding prompts, the interviewer provided extensive verbal guidance, walking participants through its intended use. As REAL ML evolved to include more guiding prompts (labeled as ``Tool'' in Table~\ref{table:total-participants}, last column), the interviewer reduced the amount of guidance they provided to participants. In later interviews, when REAL ML was more robust, participants were encouraged to follow the guiding prompts on their own while ``thinking aloud'' and asking questions as they arose. In addition to observing participants' use of REAL ML, we solicited explicit feedback on what was and was not helpful for participants and how they thought REAL ML could be improved. On average, each interview lasted 45--55 minutes. The full interview protocol is in the appendix. \looseness=-1

\paragraph{Stage 2.}
After most of the stage 1 interviews were complete, we conducted interviews with six researchers who were knowledgeable about ML (e.g., use ML in their work or regularly read ML papers) and also experts in more sociotechnical fields, such as HCI, psychology, social science, responsible AI, and research ethics. Our goal in including these participants was to surface any community norms or assumptions that might have been taken for granted by ML researchers, but would stand out to ML-adjacent researchers, many of whom have subjected the field to more critical interrogation. Some of these participants had completed our recruitment form, while others were recruited through convenience sampling with the goal of ensuring coverage of different perspectives. Table \ref{table:total-participants} contains more information about these participants. \looseness=-1

During the stage 2 interviews, we asked each participant to describe their impressions of community norms around limitations, both within and outside the ML research community, as well as any challenges or success stories. We then walked them through the tool and asked them for feedback on different sections, placing emphasis on helping ML researchers articulate limitations more effectively for different audiences, such as reviewers, researchers, and practitioners. Again, on average, each interview lasted 45--55 minutes. The full interview  protocol is in the appendix.\looseness=-1

\paragraph{Stage 3.}
After the stage 1 and stage 2 interviews were complete, we conducted an online survey with four additional ML researchers. We asked these participants to use REAL ML and complete the corresponding worksheet for an ML paper they were currently working on and planning to submit for publication.  Because of the sensitivity of requesting access to information about others' unpublished work, we limited participation to ML researchers from our own institutions and their immediate collaborators. Participants were recruited via internal mailing lists and direct emails. We sent a copy of REAL ML to each participant via email and asked them to use it on their own, without any additional guidance. After using REAL ML, each participant provided feedback via an online survey and shared their completed worksheet and limitations section with us. The feedback from participants in this stage led to very minimal requests for design changes. These requests influenced our last development iteration, resulting in the final version of REAL ML included in the appendix.\looseness=-1

\paragraph{Thematic analysis and iterative design.}
Throughout our study, we conducted a version of thematic analysis \cite{clarke2014thematic} using open coding. We first coded participants' responses and feedback into a few high level categories (e.g., general challenges, types of limitations, tool needs). We then iterated on REAL ML as new challenges and needs were identified by participants, always showing participants the latest version. Sufficient saturation was achieved on the themes before the stage 1 interviews were complete, with relatively little new information collected during later interviews. After all interviews were complete, we did one last open-coding pass on all of the responses and feedback from participants and came up with a final set of themes (e.g., fear of paper rejection, the double-edged sword of limitations, and detrimental community norms). We discuss these final themes throughout the remaining sections of this paper. \looseness=-1

\section{Perceptions of Limitations and Challenges Faced by ML Researchers}

In this section, we summarize ML researchers' perceptions of limitations, as well as the challenges they face when recognizing, exploring, and articulating limitations, thereby answering the first three research questions in Section~\ref{sec:methods}.

\subsection{Defining Limitations of ML Research}

We begin with our first research question, which asks, ``What is a limitation of ML research?'' In general, participants did not agree upon a single definition of limitations of ML research and suggested many types of limitations.\looseness=-1

\subsubsection{Limitations as inherent vs. as indicative of bad research.}
\label{sec:bad-research}

Participants had different views on limitations. Many participants took the position that limitations occur in all ML research, no matter how scientifically rigorous. P3 said, \emph{``a limitation is... something in the [methodology] that might cause us to put an asterisk on our conclusion... Like if we are saying model A is better than model B or if we are saying that you know our automated thing is comparable in performance to a human at doing a certain thing... anything that would necessitate an asterisk on that conclusion.''}  P25, an expert in social science research and ML, defined a limitation as \emph{``anything that isn't perfect.''} They called this an \emph{``extreme, but fitting''} definition, as it (accurately, in their opinion) implies that limitations are unavoidable.  Similarly, P22 indicated that in ML theory, limitations are always present, inherent in the assumptions made by researchers, stating, \emph{``from a theoretical perspective, a limitation is precisely defined as `when the assumptions are removed, your stuff doesn't work anymore.'''} This participant went on to describe the value of disclosing and discussing limitations as a reflection exercise:
\begin{quote}
\emph{``To me, limitations are an exercise of self reflection. It's a section of the research paper where authors should have a frank conversation and discussion with the readers and make them aware of some of the challenges they faced in the study. It should not be a list of things the researchers could have done or want to do in the future, it is not just admission of error... to me, it should be a `zoom-in, zoom-out' process, where the zooming-in includes taking a deeper dive into the internal and external validity of the research, acknowledging some of the assumptions and how strong these assumptions are behind a particular model. And then by zooming out, I think the limitations section should include critical thinking on the research question and how does it fit into the broader academic context, look into its policy and real-world application, as well as the potential impact on individuals, organizations, and society as a whole.''}  (P22)
\end{quote}

In contrast, other participants suggested that limitations were indicative of bad research, rather than being an inherent part of the ML research process. As one example, P19 defined limitations as \emph{``vulnerabilities''} in the ML research process, and indicated that stating limitations exposes aspects of one's research design or execution that could invite criticism. This was a common sentiment among participants, reflecting an understanding of limitations as ``weaknesses'' \cite{ross2019limited} or ``flaws'' \cite{lingard2021art} that may serve as reasons to dismiss research and the resulting findings and claims. Participants who took this position were generally unenthusiastic about highlighting these aspects of their research.

\subsubsection{Types of limitations identified by participants}

Rather than providing a definition of limitations of ML research, some participants narrowed in on a single type of limitation, often drawing on examples they had encountered previously.  Most strikingly, before seeing REAL ML, many participants defined limitations as a lack of either generalizability or robustness.  For example, P20 defined a limitation as an \emph{``attempt to draw conclusions outside of the context in which they are merited.''}
P26 instead narrowed in on technosolutionism, suggesting that ML research often involves building technologies that are not an appropriate way to address the motivating problem, a common practice that stems from the assumption that complex problems, including complex societal problems, are \emph{``solvable or computable.''} \looseness=-1

Some participants expressed a distinction between limitations that are somehow fundamental (e.g., the types of questions a model can be used to answer), limitations that arise from explicit decisions made by researchers (e.g., which hyperparameter values to use), and other limitations that arise from forces outside researchers' control (e.g., constraints on time or resources, experimental failures, or negative results). P23 referred to some limitations as \emph{``future work''} limitations---that is, things researchers did not scope or test for, but could be explored in the future. \looseness=-1

Many participants used the terms ``knowns'' and ``unknowns.'' P21, an expert in research ethics, said
they want to see \emph{``what they know, what they don't know, and what they've explored''} when reading ML papers.
Several participants brought up limitations that arise from explicit decisions made by researchers as examples of ``known knowns.''
For example, P1 explained that after testing multiple stopping criteria, they chose one that lowered accuracy but maximized efficiency. They described this as a known known because it involved a tradeoff that existed because of an explicit decision they had made.  \looseness=-1

Other participants argued that limitations should cover things researchers know they haven't explicitly tested for or measured---that is, ``known unknowns.''\footnote{Participants were not consistent in their terminology here. Some used the term ``unknown unknowns'' to describe things researchers haven't explicitly tested for or measured and can therefore only speculate about. We avoid this usage since even speculation is not possible for true unknown unknowns---that is, things researchers do not even know are possibilities.}
P4 gave an example in which they had chosen to evaluate their model using a music dataset that only included Western music, which
they knew limited the generalizability of their research claims.
Some participants drew analogies between this type of limitation and broader impacts, stating that it is important for ML researchers to think critically about the potential impacts (intended or not) of their research. P25 described known unknowns as a researcher's \emph{``I don't knows''} and argued that it was important to appropriately articulate them: \emph{``Let's hear the `I don't knows.' Let's make the `I don't knows' very explicit, and the conditions under which I do know and the conditions under which I don't know.''}  In contrast, several participants strongly opposed this idea. For example, P20 felt that speculative warnings could hinder scientific innovation because they might discourage future work or even cause harms: \emph{``I think about [speculating about known unknowns] with product warnings and health warnings and things like that. There have been a lot of myths that have endured for like 75 years, like `pregnant women shouldn't do this,' and no one ever actually empirically figured that out. And then when they do they were like `oh yeah there was no risk all along.' We were just assuming there might be and we were putting a lot of people under duress for no reason.''}\looseness=-1

\subsection{Challenges in recognizing limitations}
\label{sec:recchallenges}

One clear theme from our interviews was that junior researchers expressed difficulties recognizing the limitations of their research considerably more than senior researchers. This is partly due to a lack of transparency within the ML research community about common limitations.  Several participants said it can take junior researchers years to gain the disciplinary knowledge necessary to fully recognize the limitations of their research, in part because there are few peer-reviewed ML papers that discuss limitations, meaning their only option is to learn from their advisors or mentors over time. Some junior researchers even described how the more they participate in the ML research process, the more they realize they don't know. As P3 explained, \emph{``it's harder to identify limitations that you aren't aware of.''}\looseness=-1

Some junior researchers said they had a hard time articulating limitations without unnecessarily overemphasizing them. Both P1 and P6 said they had previously been told by their advisors that they had called out too many limitations of their research. Some junior researchers attributed their overemphasis of limitations to a lack of confidence in their research stemming from their underrepresented identities (e.g., being a woman in a field dominated by men or having English as a second language in a field in which English is the default language of publication). P24, for example, described their experience attempting to recognize limitations as a non-native English speaker as a \emph{``constant struggle.''} \looseness=-1

Although most participants who had more than five years of experience with ML research said they were confident in their abilities to recognize limitations, this confidence did not always translate to a holistic view of possible limitations. As an example, when asked to define limitations, two participants who said they did not face challenges when recognizing the limitations of their research, each with more than five years of experience, gave only limited descriptions like \emph{``lack of generalizability''} or \emph{``lack of robustness.''} It is therefore possible that senior researchers could still benefit from increased support recognizing limitations even if they feel confident doing so on their own. \looseness=-1

\subsection{Challenges in exploring and articulating limitations}
\label{sec:challengesEandA}

Participants expressed a variety of challenges related to exploring and articulating limitations. Some challenges, like fear of paper rejection, echo those observed in other fields~\cite{puhan2012discussing,brutus2013self,ross2019limited,lingard2021art}, while others are more specific to community norms.\looseness=-1

\subsubsection{Limitations as grounds for rejection.}
\label{sec:rejection}

One theme touched on by over half of the participants was the fear that disclosing limitations would increase the likelihood of paper rejection. As P2 explained, limitations are a double-edged sword, since disclosing them can make research appear less scientifically rigorous, but omitting them can suggest that the researchers do not understand the implications of their research design or execution: \emph{``not mentioning enough limitations is grounds for rejection, but mentioning too many limitations is also grounds for rejection,''} (P2). This participant went on to say they tried to disclose only a few limitations in their papers, even if they knew of other limitations, because they did not want to give \emph{``fuel''} to reviewers to \emph{``trash their paper.''}  P15 said, on average, they would include only 3--4 limitations per paper, in part due to pragmatic reasons, but also \emph{``to not admit to too many limitations for the reviewers.''}  Similarly, P14 mentioned that even though they thought it was important to disclose limitations, they were hesitant to do so because a long list of limitations might undermine the perceived importance or benefits of their research and \emph{``you don't want to shoot yourself in the foot when you are writing a limitations section.''} \looseness=-1

Relatedly, some participants decided to omit limitations due to the perceived stigma around disclosing them, as discussed in Section~\ref{sec:bad-research}. P11 admitted that some of the limitations of their research were important due to their potential negative impacts on society, but chose not to disclose them \emph{``because societal impacts aren't mentioned in other ML research papers,''} going on to say they would not feel comfortable disclosing them because of community norms. In general, the choice to disclose less information about limitations appeared to be primarily motivated by a desire to appease reviewers.\looseness=-1

Concerns about paper rejection may be compounded by perceived differences in reviewers' expectations of which limitations are important to disclose---and, more generally, which values are important to prioritize---which can also vary by subfield and by publication venue. P2 shared the example that reviewers for robotics publication venues tend to prioritize practicality over novelty or theoretical claims, while reviewers for other ML publication venues may not. As P5 put it, \emph{``as you are going through this process, you understand what certain conferences are looking for.'''}\looseness=-1

Multiple participants confessed that the emphasis on abstraction and generalizability in the ML review process had led them to underemphasize or omit some of the limitations of their research, since limitations often relate to the ways in which research cannot or should not be generalized. As P16 explained, \emph{``work that is generalizable is more valued, so in research you want to overemphasize your generalizability---even if it isn't necessarily true.''}\looseness=-1

\subsubsection{Struggles with prioritization and organization.}

Another commonly expressed challenge involved the page limits imposed by ML publication venues. P7 said, \emph{``usually authors are constrained to be very precise about their limitations, and it can be hard to compress all of that information into the page limits that they have.''}  Given limited space, participants felt the need to prioritize limitations, and several participants mentioned not knowing how to do this.  For example, is it best to narrow in on the limitation the researchers personally find most important and discuss it in depth, or to cover more limitations at a higher level?  P21 suggested that \emph{``in the event of page limits... authors should prioritize the limitations with the highest known severity of impact,''} but severity is not always easy to judge.  Indeed, P23, who had previously done work in research ethics, mentioned that the ML researchers they worked with consistently requested a ``Richter scale'' of some kind to measure the severity of the impacts of limitations so they would know which to articulate first.\looseness=-1

Participants also struggled with where and how to articulate limitations.  Some participants had difficulty knowing how to develop a narrative around limitations that would be valuable to different audiences. As P6 said, \emph{``the most difficult thing for me is how to write a [...] coherent story.''}  Several other participants said they were not sure if this ``\emph{story}'' should appear in a dedicated limitations section or if limitations should be introduced throughout the paper. Finally, several participants said they were unsure how much information was too much. They noted that some types of limitations occur so commonly in ML research that discussing such limitations would risk wasting readers' time. P12 said they struggled with knowing where to draw the line when articulating limitations, noting that \emph{``it's hard to [strike] the balance between simply listing some caveats and it turning into a total philosophical paper.''}  Similarly, P4 said they could have included an extra two pages in their paper on their dataset's lack of generalizability across cultures and the societal impacts of this limitation, but thought it \emph{``would probably be a waste of time for the ML research audience.''}\looseness=-1

\section{REAL ML}
\label{sec:tool}
\begin{figure}[ht]
\centering
\includegraphics[width=\linewidth]{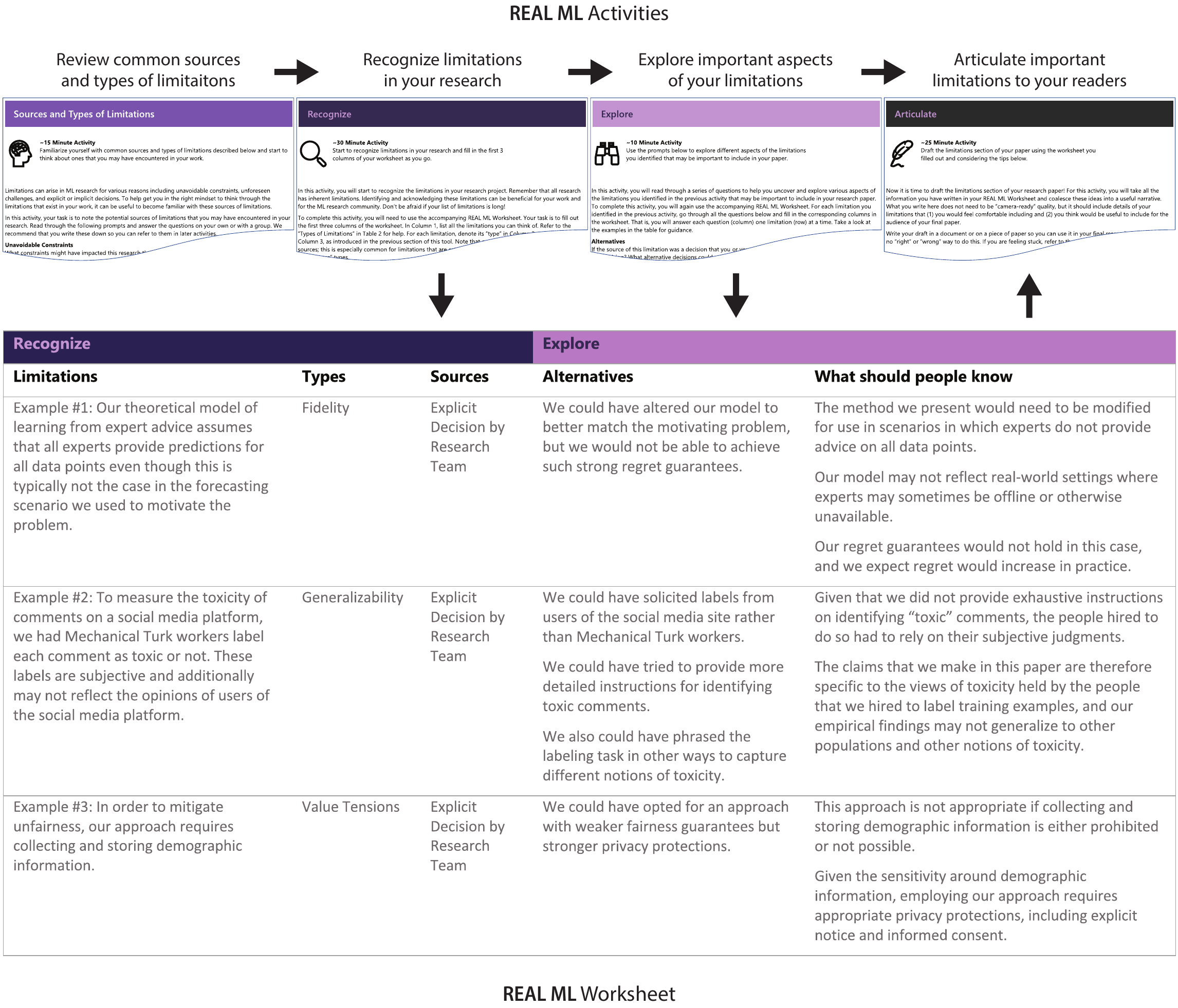}
  \caption{An overview of the instructions, activities, and worksheet included in REAL ML.}
  \Description{A flowchart of the REAL ML workflow, showcasing the four categories of activities in the tool. The tool starts with a review of common sources and types of limitations. It then includes sections on recognizing limitations in your current research, exploring important aspects of these limitations, and finally articulating key limitations to your readers. The recognize and explore sections of REAL ML feed into the REAL ML Worksheet, where each row is a limitation and columns include a description of the limitation, its type, its source, alternatives, and what people should know about the limitation. The worksheet feeds into to the final section of the tool (articulate).}
  \label{fig:overview}
\end{figure}

Our findings reveal many challenges faced by ML researchers. Some of these challenges stem from community norms that disincentivize disclosure and discussion of limitations, but other challenges can be attributed to a lack of guidance and appropriate training on recognizing, exploring, and articulating limitations. When developing REAL ML, we therefore took into account participants' expressed needs for guidance, iteratively updating the tool based on their feedback and on our observations of their attempts to use it. In this section, we describe REAL ML and the reasoning behind our main design decisions. Figure~\ref{fig:overview} provides an overview of REAL ML, which is broken down into an introduction plus four content sections: 1) sources and types of limitations, 2) recognizing limitations, 3) exploring limitations, and 4) articulating limitations. Each section includes guided activities and resources for ML researchers to use when writing limitations sections. We describe each section below; the full tool is in the appendix and available at \url{https://github.com/jesmith14/REAL-ML}.\looseness=-1

\subsection{Sources and Types of Limitations}

The first section asks ML researchers to familiarize themselves with sources of limitations and types of limitations that occur commonly in ML research and to start thinking about how these might relate to their research.  Sources of limitations are broken into three broad categories: unavoidable constraints (e.g., constraints on time or resources), unforeseen challenges (e.g., experimental failures or negative results), and implicit and explicit decisions made during the ML research process. For this last category, REAL ML directs ML researchers to a list of common decision-making points in the ML research process where limitations could arise, as shown in Table \ref{table:decision-making-points}. They are prompted to use this guidance to reflect on possible sources of limitations of their research.  Finally, they are presented with a list of types of limitations that occur commonly in ML research, as shown in Table \ref{table:limitation-types}, to reflect on and return to later.

\begin{table*}[t!]
  \caption{Decision-making points presented in the ``sources and types of limitations'' section of REAL ML.}
  \label{table:decision-making-points}
\begin{tabular}{p{3.6cm}p{10.4cm}}
\toprule
                \textbf{Decision-making points} &                                         \textbf{Examples} \\
\midrule
        Composition of the research team & Demographic features (e.g., race, gender), disciplinary training (e.g., computer science, medicine), epistemological perspectives (e.g., Bayesian vs. Frequentist), or other researcher characteristics that can influence the approach to research and interpretation of the findings \\
        \hline
              Related work & The specific fields with which your current study is engaging and which may shape your research; prior work to which your current work is responding; prior work upon which your current work is building \\
              \hline
                    Problem formulation & The general problem that motivates the research; the specific research questions developed to get at that problem \\
                    \hline
     Formalism of the problem & Mathematical statement of the problem that your study is trying to address; technical assumptions (e.g., i.i.d. data points) \\
     \hline
                            Technical approach & Learning algorithm; statistical model; hyperparameter choices \\
                            \hline
        Theoretical claims & Theoretical guarantees such as error bounds; analyses of computational complexity; mathematical derivations \\
        \hline
        Datasets & The collection, curation, and selection of datasets; the use of particular datasets for training or evaluating \\
        \hline
        Empirical evaluation setup and metrics & Experimental setup including approaches to be compared, metrics, parameter settings; research subjects \\
        \hline
        Ablation studies & Setup for ablation studies, including components removed and metrics \\

\bottomrule
\end{tabular}
\end{table*}

\begin{table*}[t!]
  \caption{Types of limitations presented in REAL ML. (The tool includes additional examples of each type of limitation.)}
  \label{table:limitation-types}
\begin{tabular}{p{2.6cm}p{6.3cm}p{5.1cm}}
\toprule
                \textbf{Types \linebreak of Limitations} &                       \textbf{Probes to Uncover Limitation} & \textbf{Examples} \\
\midrule
                            Fidelity & How faithfully do the formalism of the problem, the technical approach, and the results map onto the motivating problem that drives the work? & The training data was labeled even though similar real-world data is not usually labeled. \\
                            \hline
                    Generalizability & To what extent do the results hold in different contexts? How broadly or narrowly should the claims in the paper be interpreted? How broadly can the technical approach be applied across domains? & Model was developed for a particular scenario and does not apply to other scenarios or contexts. \\
                    \hline
                          Robustness & How sensitive are the results to minor violations of assumptions (e.g., small tweaks to mathematical model, metrics, hyperparameters)? & Adding a small amount of noise in the data dramatically reduces accuracy. \\
                          \hline
                       Reproducibility & To what extent could other researchers reproduce the study? & Researchers provide details on parameter settings used but cannot share code or data because they are proprietary. \\
                       \hline
               Resource \linebreak Requirements & Is the technical approach computationally efficient? Does it scale? What other resources does the technical approach require? & Technical approach requires specialized hardware. \\
               \hline
                      Value Tensions & Are some values (e.g., novelty, simplicity, high accuracy, low false positive rate, ease of implementation, interpretability, efficiency) sacrificed in pursuit of others? & The model has high accuracy on a test dataset but is a black box and hard to interpret. \\
                      \hline
                Vulnerability to Mistakes and Misuse & How sensitive are the results to human errors, unintended uses, or malicious uses? & System operators are liable to misinterpret results without sufficient training. \\
\bottomrule
\end{tabular}
\end{table*}

All versions of REAL ML, starting with the initial prototype, included something akin to Tables~\ref{table:decision-making-points} and \ref{table:limitation-types}, but the content and placement evolved over the course of our study. Some decision-making points (e.g., the composition of the research team and ablation studies) were suggested by participants. Participants who saw early versions of REAL ML also pointed out that limitations can arise from sources other than decisions made by researchers. For example, P2 noted that one of their biggest sources of limitations was \emph{``time constraints.''}
P5 mentioned that limitations were sometimes caused by experimental failures. They said, as a reader of ML papers, they often wished they knew about limitations arising from other ML researchers' failed experiments so they would know what not to do in the future.
P7 suggested that limitations fall into two categories: those that researchers have control over and those that they do not. They went on to explain that \emph{``[lack of access to compute] is a type of limitation that I have no control over, I can do my best... but I can [only] do whatever I can with the small amount of compute that I have, so this is a limitation that is out of my control.''}
Our decision to include unavoidable constraints and unforeseen challenges as sources of limitations was based on this early feedback from participants.\looseness=-1

Initially there was disagreement among our team as to whether it would be more effective to ask ML researchers to begin by reflecting on possible sources of limitations of their research before considering types of limitations or to instead begin with types of limitations before considering sources of limitations.  After trying both approaches with different participants, we found that beginning with sources of limitations was a more successful way to spark open-ended brainstorming.  This approach gave participants an opportunity to reflect on why they had made various decisions, which may have led to limitations, during the ML research process. In contrast, jumping straight to types of limitations without this initial reflection caused participants to anchor too much on the examples we provided. \looseness=-1

\subsection{Recognizing Limitations}

The next section prompts ML researchers to build on the brainstorming activity in the first section by filling in a worksheet with a list of the limitations of their research, along with their sources and types. The worksheet is in the appendix. \looseness=-1

This section evolved conjointly with the previous section. We found that participants were able to more easily recognize the limitations of their research after brainstorming about sources and types of limitations. For example, P22 was pleasantly surprised by their ability to recognize new limitations of their research after being exposed to Table \ref{table:limitation-types}, and later said, ``\emph{I can honestly see all of these limitation types applying to my work---and honestly to all ML work.}''  As discussed in Section~\ref{sec:recchallenges}, junior researchers were particularly eager for guidance about recognizing limitations, and of the four survey participants, we found that the one junior researcher rated this section as more useful than the three senior researchers.\looseness=-1

\subsection{Exploring Limitations}
\label{sec:explore}

This section asks ML researchers to answer a series of questions designed to help them explore the limitations they recognized in the previous section, with the goal of uncovering information that may be important to articulate. They are asked to record their answers in the worksheet. For each limitation, they are first asked to think through potential alternative approaches that could have been taken---that is, alternative decisions that could have been made or alternative research designs that could have been explored---as well as pros and cons of each. Next, they are are prompted to reflect on what different audiences might need to know about this limitation, considering the distinct needs of reviewers, researchers, practitioners, and people who use or are affected by ML systems that build on their research.\looseness=-1

This section was designed to address challenges around determining which information to focus on. We took the approach of encouraging ML researchers to think through the information that would be most valuable to different audiences, rather than dictating what we felt was most important.  In early versions of REAL ML, we included probing questions drawn directly from the four-step process of \citet{ross2019limited}, asking about potential alternative approaches, the potential impacts of each limitation, and how these impacts were mitigated. Based on participant feedback, we later deemphasized mitigating impacts so that researchers who had not already taken explicit steps to mitigate impacts would still have an opportunity to reflect on how these impacts could be mitigated in the future.  Finally, three of the four survey participants indicated that listing impacts of limitations in the worksheet felt redundant with other questions, leading us to merge the consideration of impacts into the reflection about what different audiences might need to know.\looseness=-1

\subsection{Articulating Limitations}

In the final section, ML researchers are asked to build on the information they recorded in the worksheet to draft a limitations section.  The goal is to develop a narrative around the limitations of their research that is valuable to different audiences.  The narrative does not need to be of ``camera-ready'' quality and, indeed, they may choose to introduce information about limitations throughout their paper rather than including it in a dedicated limitations section. To help with narrative development, REAL ML includes a set of ``tips and tricks'' for articulating limitations, including guidance about speculation, prioritization, broader impacts of limitations, and concerns about paper rejection.\looseness=-1

The ``tips and tricks'' were targeted at addressing needs repeatedly expressed by our participants, like struggles with prioritization and organization, as discussed in Section~\ref{sec:challengesEandA}.  The specific guidance on prioritization---that is, focusing on limitations that might have the most severe impacts, in addition to those that would be most valuable for different audiences to know about---was inspired, in part, by remarks from P23, who mentioned the need for a ``Richter scale'' to measure the severity of the impacts of limitations. However, in order to avoid embedding our own biases about impacts into REAL ML, we opted to leave the determination of severity to ML researchers using the tool.\looseness=-1

All four survey participants said they ``agreed'' or ``strongly agreed'' with the statement ``I felt better prepared to write about these limitations than I would have been without the tool,'' an indication that---at least for this very small set of participants---even senior researchers who felt comfortable recognizing the limitations of their research found the tool useful for articulating those limitations, although a larger evaluation study would be needed to fully measure the benefits.\looseness=-1

\section{Limitations, Discussion, and Future Work}
\label{sec:discussion}

Our study itself has limitations that could impact both our research findings and REAL ML. First, we conducted our qualitative research through an interpretivist lens. As such, our research findings reflect our own biases and subjectivities. Second, because we largely recruited participants via our personal networks, selection bias was unavoidable. Notably, we ended up with an imbalanced sample in terms of research backgrounds. For example, many interview participants had expertise in natural language processing, but only one had expertise in learning theory. It is also possible that participants were more open to recognizing, exploring, and articulating the limitations of their research than is typical of ML researchers. These limitations may have impacted both the perspectives reflected in our research findings and the evolution of REAL ML. ML researchers who use REAL ML should recognize they may have different needs when it comes to articulating the limitations of their research (e.g., preferring to introduce information about limitations throughout the paper rather than including it in a dedicated limitations section, recognizing types of limitations other than the ones listed in REAL ML). REAL ML is meant to act as a guide for ML researchers. It is not all-encompassing nor prescriptive, and we recommend that ML researchers adapt both the suggested activities and the outputs to meet their needs.  (Although this paper is not a typical ML paper and therefore not the type of paper that REAL ML was intended for, we adapted the suggested activities to guide us when writing this limitations section.)
Additionally, since our study focused on iteratively developing and testing REAL ML rather than empirically evaluating it in action, we have not yet conducted a full evaluation of the final version with ML researchers beyond the four survey participants. Additional research is needed to measure the effectiveness of REAL ML in practice.\looseness=-1

As discussed in Section~\ref{sec:intro}, REAL ML is intended for use \emph{posthoc} via a process known as ``reflection on action'' \cite{munby1989reflection}, helping ML researchers look back on their research during the paper-writing stage of the ML research process. Several interview participants said they would find it valuable to have a similar tool targeted at earlier stages of the ML research process, noting that if they had thought through the limitations of their research earlier, they would have pivoted their research direction or made different decisions.  A natural next step would therefore be to adapt REAL ML to explicitly target earlier stages of the ML research process in order to encourage ``reflection in action'' \cite{schon1984architectural}.\looseness=-1

REAL ML was developed to address the challenges faced by ML researchers when recognizing, exploring, and articulating limitations.  However, we note that some of these challenges go beyond the scope of what can be accomplished with a tool, requiring broader shifts in community norms to address.
For example, many participants expressed concerns about the emphasis on generalizability in the ML review process, mentioning that this had led them to underemphasize or omit some of the limitations of their research. As others have noted \cite{d2020data,birhane2021values}, generalizability is highly valued in the ML research community. Deemphasizing generalizability would require a major shift in community norms.
Along similar lines, as discussed in Section~\ref{sec:rejection}, many participants expressed concerns about whether disclosing limitations would increase the likelihood of paper rejection, in part because of ML publication venues' reviewing norms and in part because of perceived differences in reviewers' opinions about disclosing limitations. Participants suggested that a version of REAL ML be provided to reviewers in order to standardize the way limitations are critiqued during the ML review process.
Standardized guidance for reviewers could serve as a complement to emerging efforts to promote transparency on the part of authors, such as the introduction of broader impacts statements at NeurIPS 2020~\cite{neurips-broaderimpact} and the subsequent NeurIPS 2021 paper checklist~\cite{neurips-paperchecklist}, which encouraged authors to disclose the limitations of their research.
Of course, more research is needed to explore how reviewers would respond to heightened transparency around limitations and whether standardizing the way limitations are critiqued would improve the ML review process.\looseness=-1

\section{Conclusion}

Despite previous work calling out the importance of transparency around limitations, the ML research community lacks well-developed norms around disclosing and discussing limitations. In this paper, we uncovered the practical and cultural challenges faced by ML researchers when recognizing, exploring, and articulating limitations. Specifically, we found that the ML research community does not have a single, agreed-upon definition of limitations of ML research, nor does it have a standardized process for disclosing and discussing limitations. We discovered that junior researchers were particularly eager for guidance about recognizing limitations, and that both junior and senior researchers would benefit from guidance about articulating limitations. Using a three-stage interview and survey study, we conducted an iterative design process to develop and test REAL ML, a set of guided activities to help ML researchers recognize, explore, and articulate the limitations of their research. REAL ML was intended to address some of the practical challenges faced by ML researchers. However, our study also exposes cultural challenges that go beyond the scope of REAL ML and will require broader shifts in community norms to address. We hope our study and REAL ML help move the ML research community toward more actively and appropriately engaging with limitations. \looseness=-1

\section*{Acknowledgments}

We are grateful to all of our study participants for their time. We also thank members of Microsoft's FATE research group, who provided valuable feedback throughout our study and the development of REAL ML. Special thanks goes to David Alvarez-Melis, Zana Bu\c{c}inca, Mahsan Nourani, Divya Shanmugam, and Angelina Wang, who volunteered their time to help us pilot our interview protocol.\looseness=-1

\section*{Funding/Support} Most of this research was conducted while the first author was an intern at Microsoft. All other authors are full-time employees of Microsoft.

\bibliographystyle{ACM-Reference-Format}
\bibliography{main}

\newpage

\appendix

\section{REAL ML}

This section includes screenshots of REAL ML. The full tool is available at \url{https://github.com/jesmith14/REAL-ML}.

\begin{figure}[ht]
    \caption{REAL ML, Page 1.}
  \centering
  \frame{\includegraphics[scale=0.30]{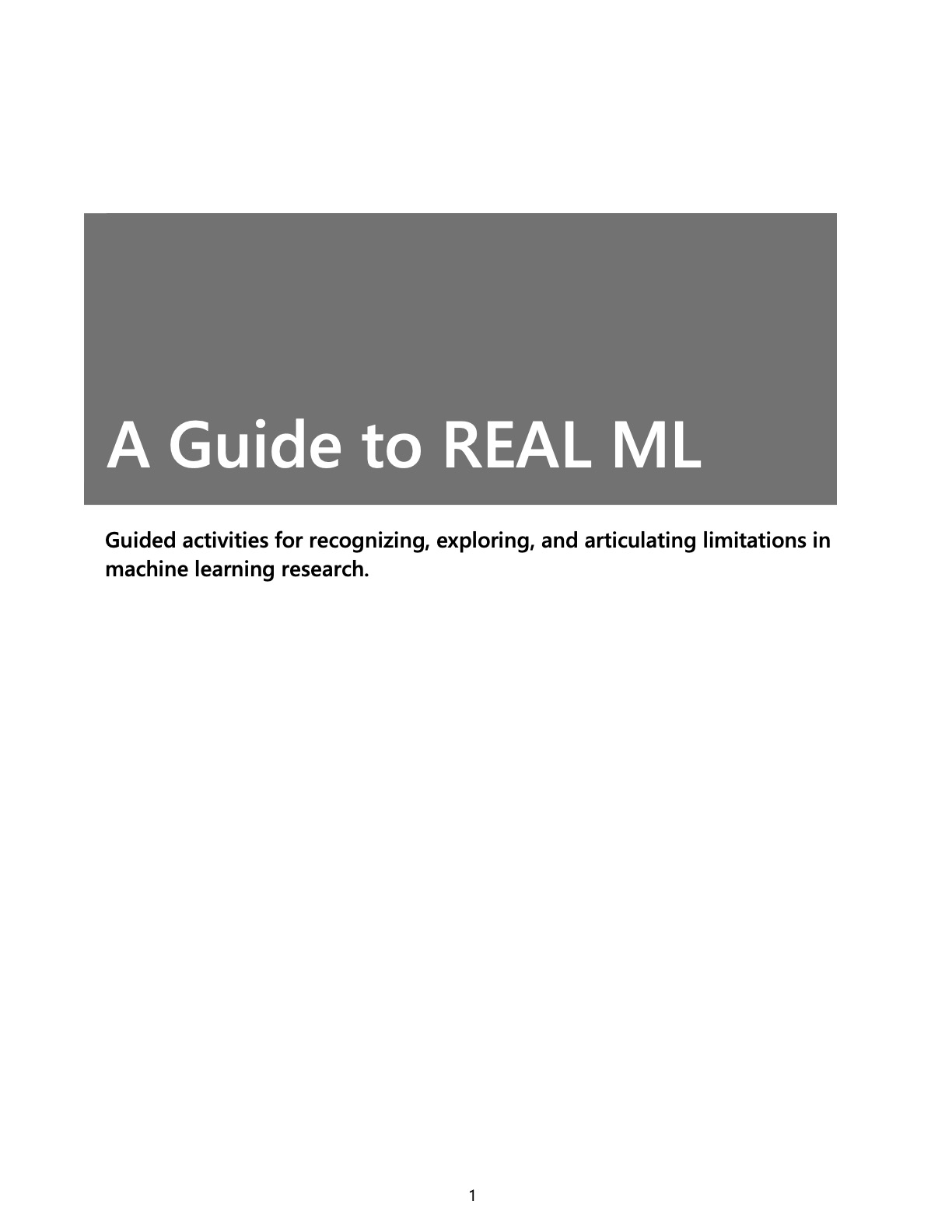}}
  \label{fig:REALMLPage1}
\end{figure}

\clearpage

\begin{figure}[ht]
  \caption{Real ML, Page 2.}
  \centering
  \frame{\includegraphics[scale=0.33]{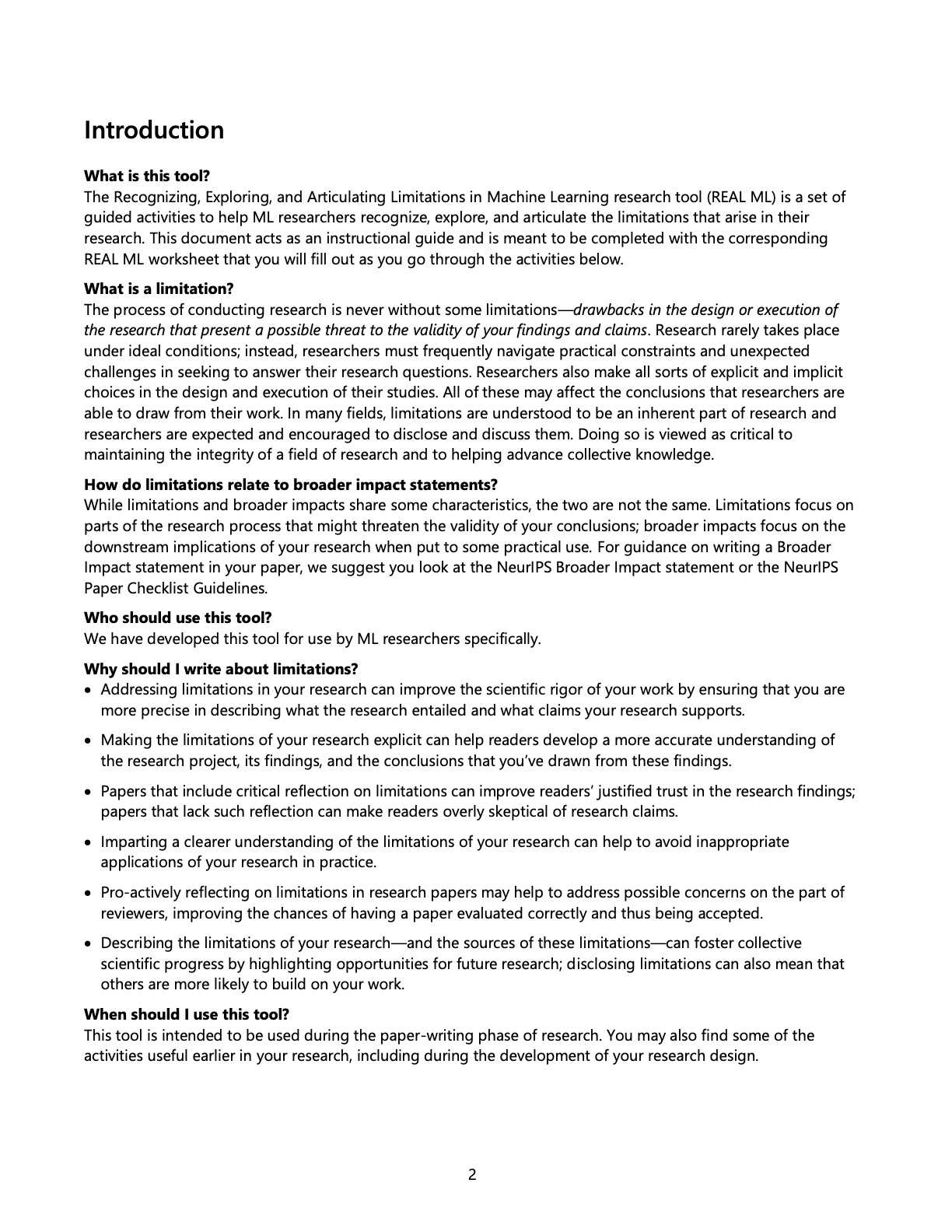}}
  \label{fig:REALMLPage2}
\end{figure}

\clearpage

\begin{figure}[ht]
  \caption{Real ML, Page 3.}
  \centering
  \frame{\includegraphics[scale=0.33]{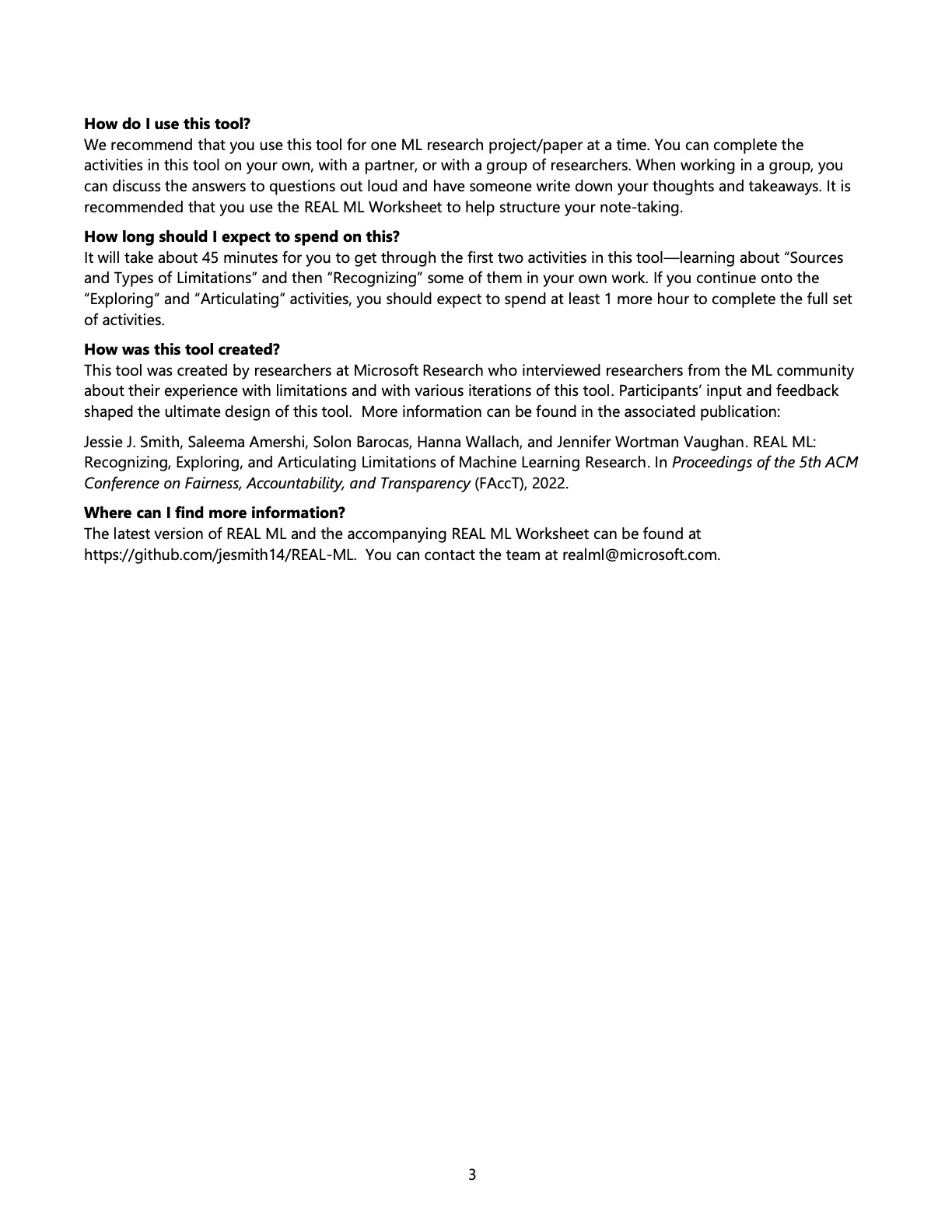}}
  \label{fig:REALMLPage3}
\end{figure}

\clearpage

\begin{figure}[ht]
  \caption{Real ML, Page 4.}
  \centering
  \frame{\includegraphics[scale=0.33]{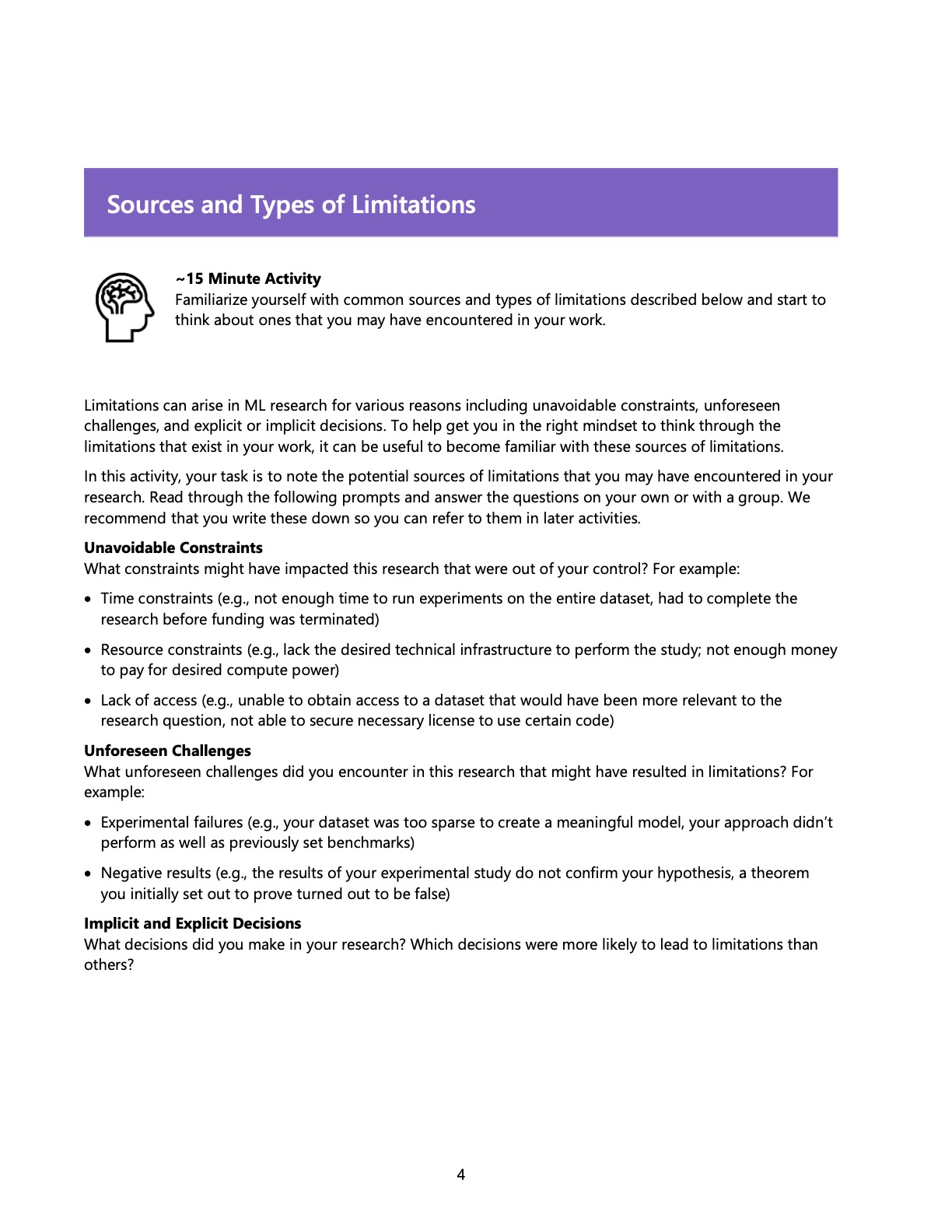}}
  \label{fig:REALMLPage4}
\end{figure}

\clearpage

\begin{figure}[ht]
  \caption{Real ML, Page 5.}
  \centering
  \frame{\includegraphics[scale=0.33]{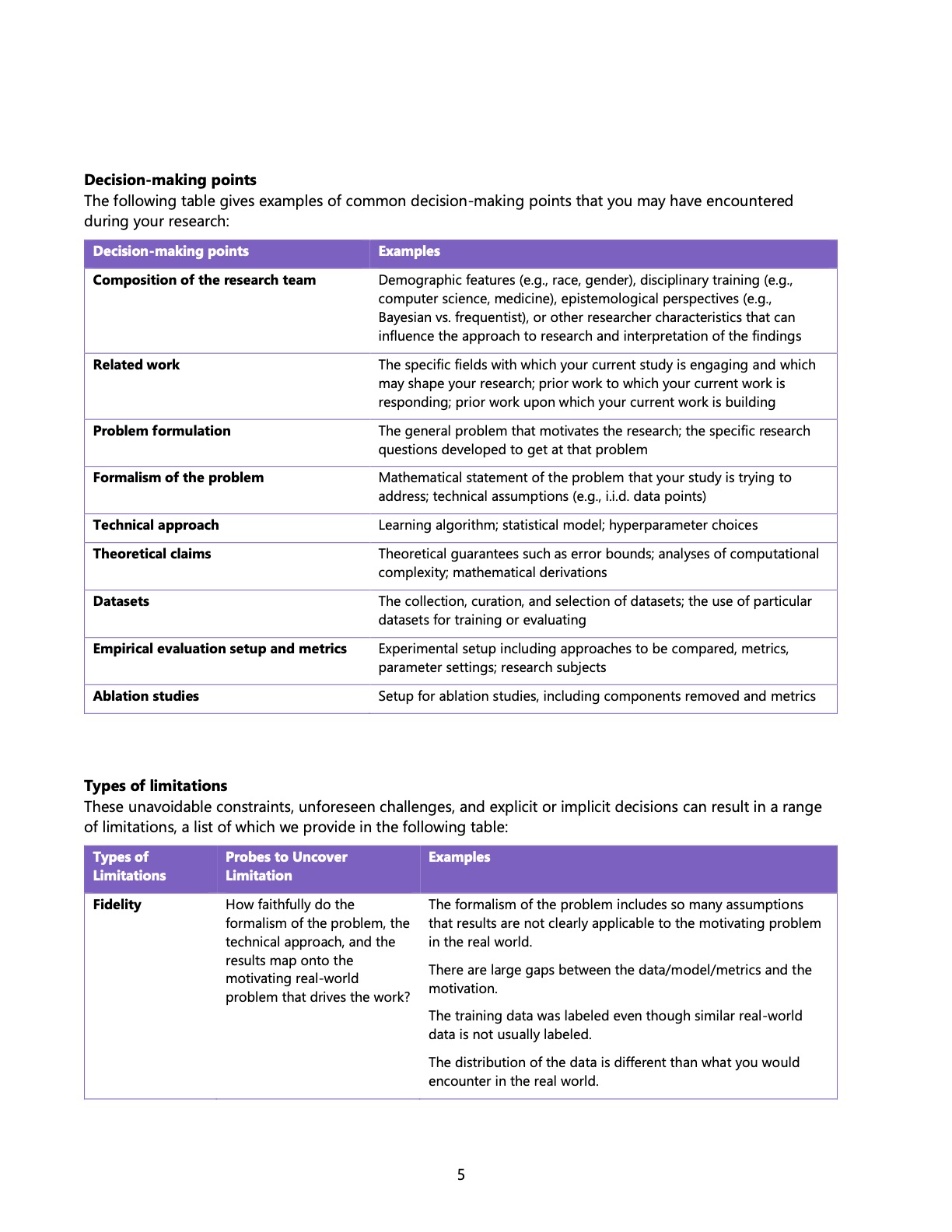}}
  \label{fig:REALMLPage5}
\end{figure}

\clearpage

\begin{figure}[ht]
  \caption{Real ML, Page 6.}
  \centering
  \frame{\includegraphics[scale=0.33]{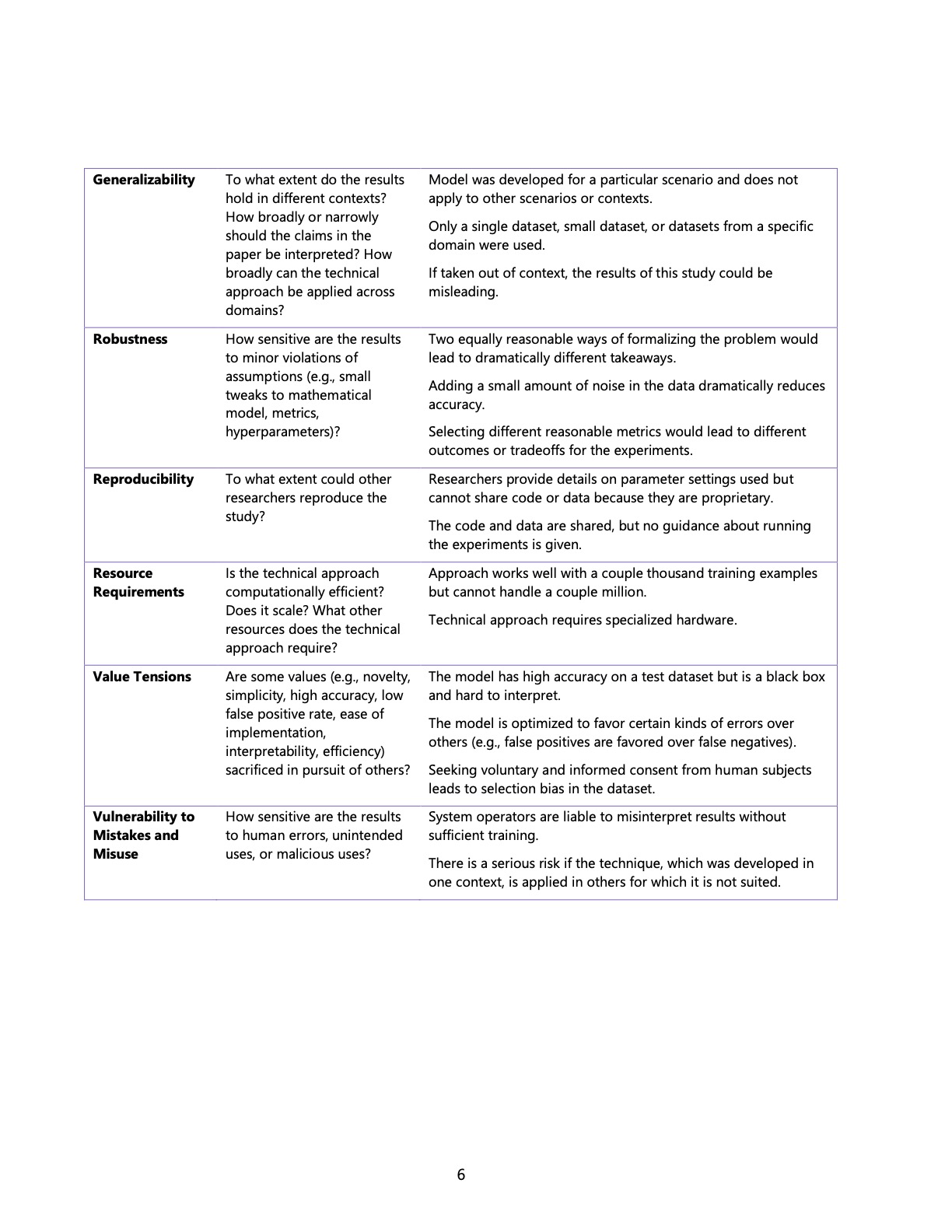}}
  \label{fig:REALMLPage6}
\end{figure}

\clearpage

\begin{figure}[ht]
  \caption{Real ML, Page 7.}
  \centering
  \frame{\includegraphics[scale=0.33]{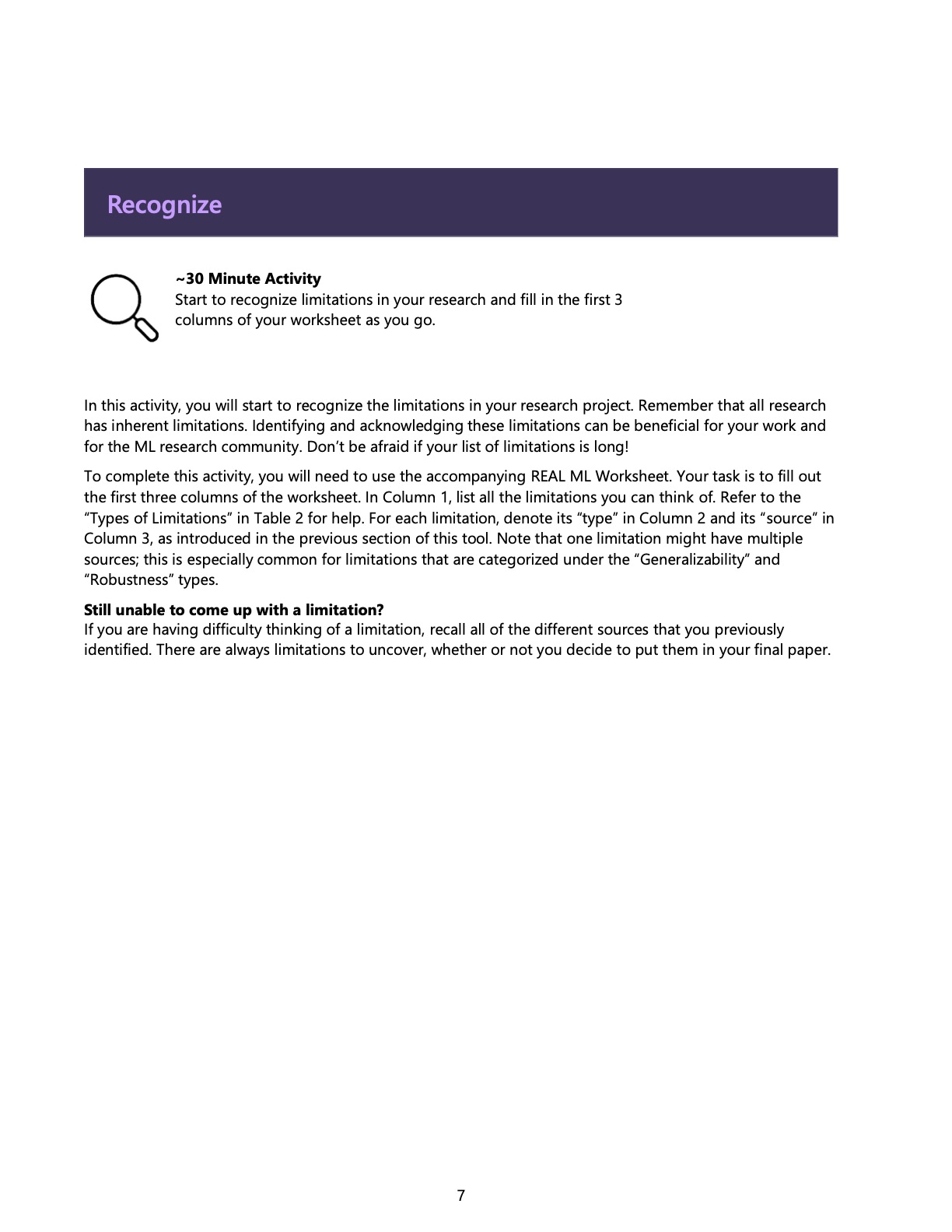}}
  \label{fig:REALMLPage7}
\end{figure}

\clearpage

\begin{figure}[ht]
  \caption{Real ML, Page 8.}
  \centering
  \frame{\includegraphics[scale=0.33]{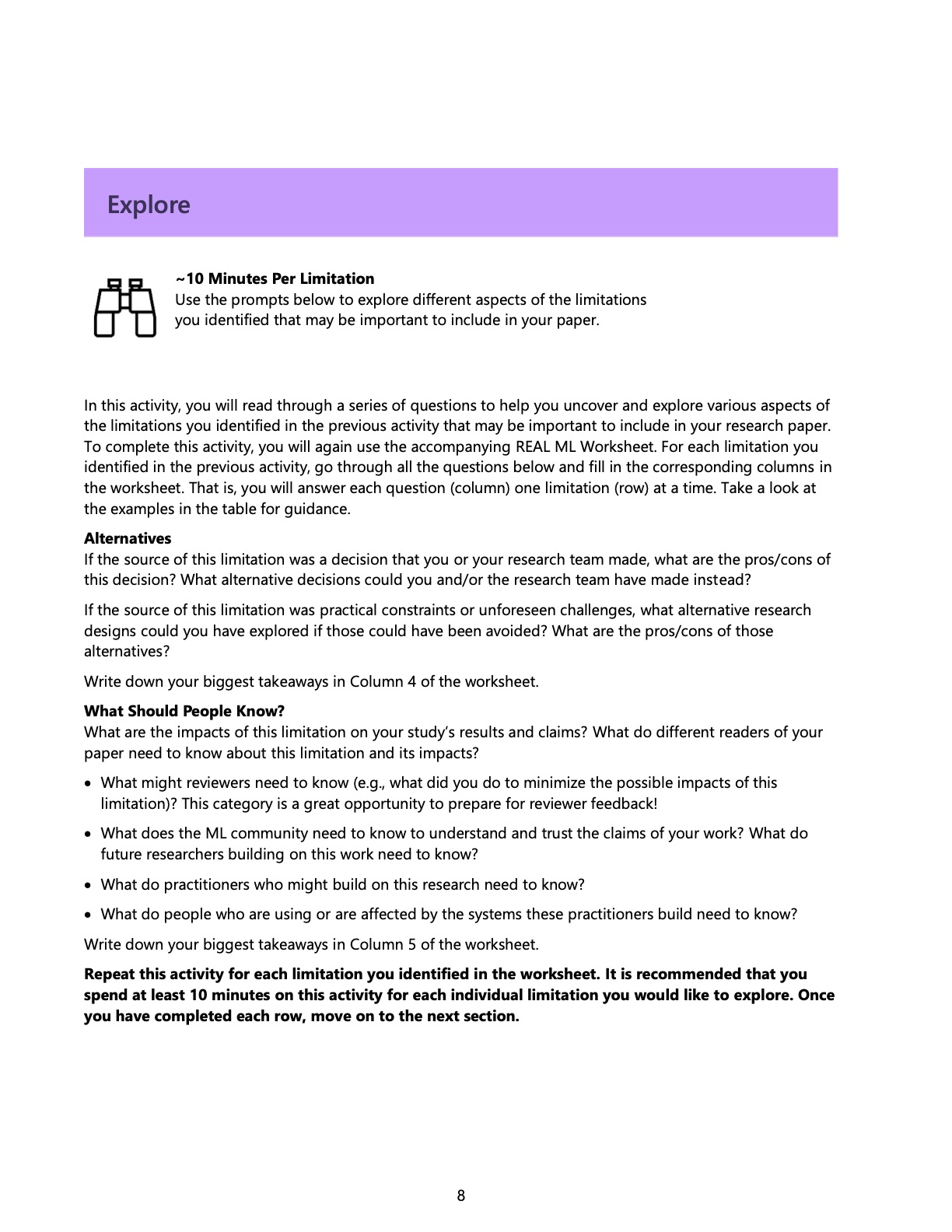}}
  \label{fig:REALMLPage8}
\end{figure}

\clearpage

\begin{figure}[ht]
    \caption{Real ML, Page 9.}
  \centering
  \frame{\includegraphics[scale=0.33]{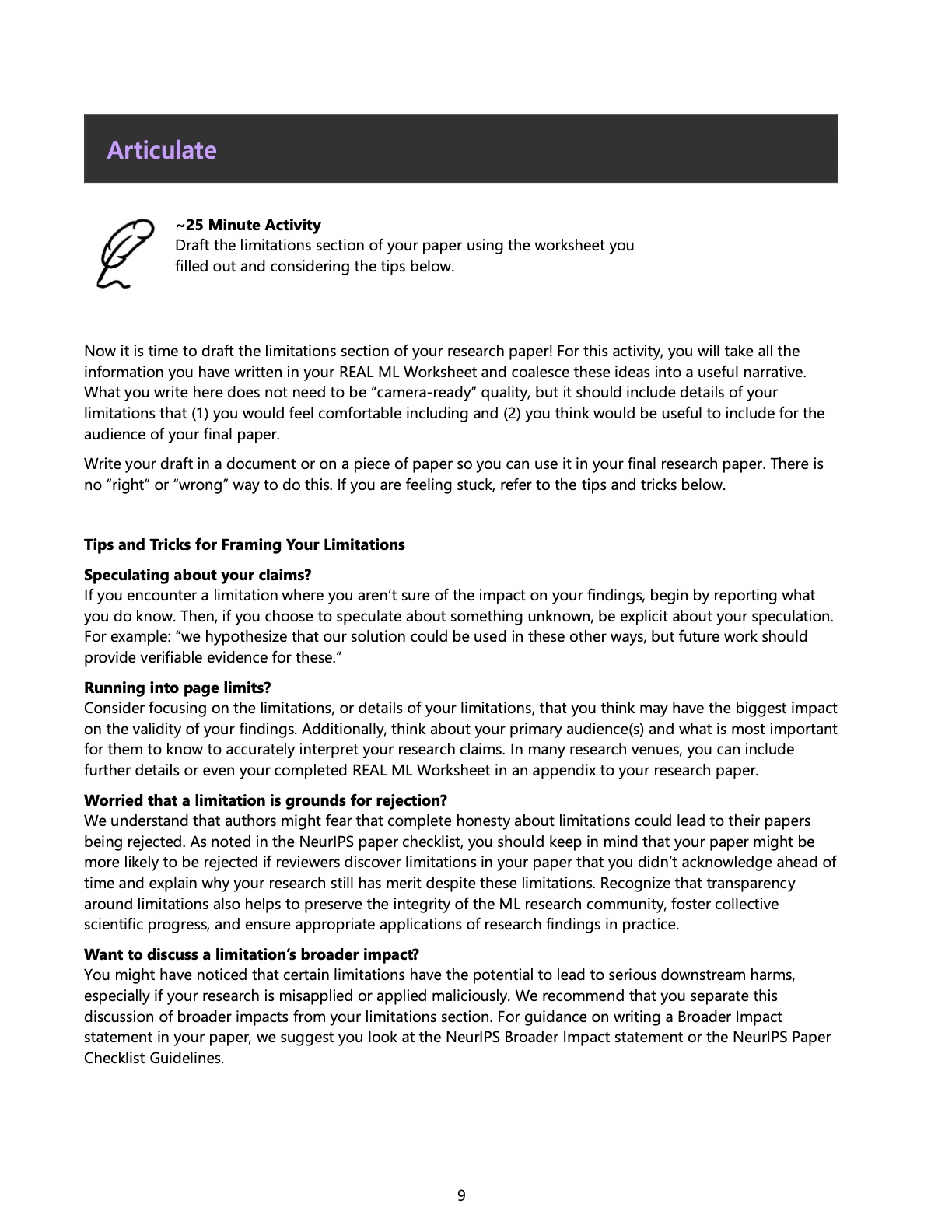}}
  \label{fig:REALMLPage9}
\end{figure}

\clearpage

\begin{figure}[ht]

  \centering
  \includegraphics[width=\linewidth]{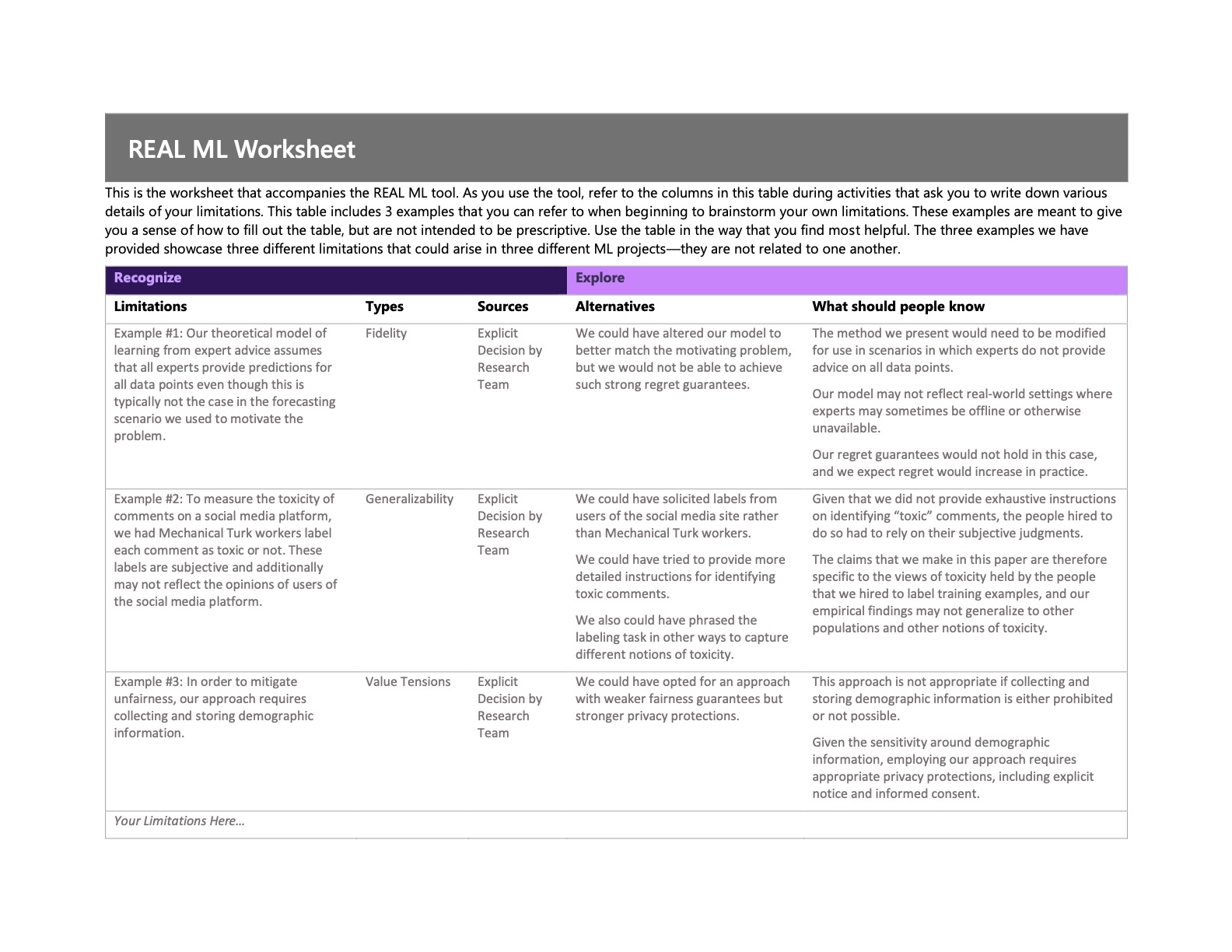}
  \caption{REAL ML, Page 10. The REAL ML Worksheet.}
  \label{fig:REALMLWorksheet}
\end{figure}

\clearpage

\section{Tool Prototype}

The initial prototype of REAL ML consisted of three lists intended to encourage reflection: a list of types of limitations that commonly occur in ML research (Table~\ref{table:prototype-limitations}), a list of common decision-making points in the ML research process where limitations could arise (Table~\ref{table:prototype-decisions}), and a list of probing questions to answer when preparing to articulate a limitation (Table~\ref{table:prototype-steps}). The versions included here correspond to the V1 prototype referenced in Table~\ref{table:total-participants}, the earliest version used in our study.\looseness=-1

\begin{table*}[ht]
  \caption{The types of limitations included in the V1 prototype of REAL ML. Through our iterative design process, this table evolved into the one included in the ``sources and types of limitations'' section of REAL ML, shown in Table~\ref{table:limitation-types} (with pared-down examples).}
  \label{table:prototype-limitations}
\begin{tabular}{p{3cm}p{4cm}p{7cm}}
\toprule
                \textbf{Types of Limitations} &                           \textbf{Definition of Limitation} &                                 \textbf{Examples \& Impacts} \\
\midrule
                            Fidelity (or Applicability) & How faithfully do the formalism of the problem, the technical approach, and the results map onto the motivating real-world problem that drives the work? & The formalism of the problem includes so many assumptions that results are not clearly applicable to the motivating problem in the real world. \\
                            \hline
                    Robustness & How sensitive are the results to (minor) violations of assumptions?  & Two equally reasonable ways of formalizing the problem lead to dramatically different takeaways. Adding a small amount of noise in the data dramatically reduces accuracy. If a system operator misinterprets results, bad things could happen. \\
                    \hline
                          Generalizability & To what extent do the results hold in different contexts? How broadly can the technical approach be applied?  & Model is developed for a particular scenario and does not apply to other scenarios or contexts, e.g., face recognition system requires good lighting. \\
                          \hline
                       Replicability & To what extent could other researchers replicate the results? & Authors provide details on parameter settings used but cannot share code or data because they are proprietary. \\
                       \hline
               Scope of Claims & How broadly or narrowly should the claims in the paper be interpreted? How do the implicit assumptions made in the evaluation impact interpretation of results? & Only a single random seed was tried. Only a single dataset was used.  Only one part of a more complex system was modeled. \\
               \hline
                      Validity/Rigor &   Were there better research practices that could have been used? & Because access to data was limited, multiple experiments were run iteratively using the same test data. \\
                      \hline
Computational Efficiency & Is the technical approach computationally efficient? Does it scale? & Approach works well with a couple thousand training examples, cannot handle a couple million. Approach requires a lot of computational resources and therefore has bad carbon footprint. \\
\hline
Other Costs & What other resources does the technical approach require? & Technical approach requires specialized hardware.\\
\hline
Other Tradeoffs & What tradeoffs must be made to achieve the results? & Technical approach has good accuracy, but sacrifices privacy.\\
\bottomrule
\end{tabular}
\end{table*}

\begin{table*}[ht]
  \caption{The decision-making points included in the V1 prototype of REAL ML. Through our iterative design process, this table evolved into the one included in the ``sources and types of limitations'' section of REAL ML, shown in Table~\ref{table:decision-making-points}.}
  \label{table:prototype-decisions}
\begin{tabular}{p{5cm}p{9cm}}
\toprule
                \textbf{Decision making points} &                                         \textbf{Definition} \\
\midrule
        High level problem formulation & High level description of the problem this paper trying to solve; motivation for the problem; research questions \\
        \hline
              Formalism of the problem & Mathematical statement of the problem this paper is trying to solve; technical assumptions (e.g., i.i.d. data points, adversarial noise) \\
              \hline
                    Technical approach & Learning algorithm; statistical model \\
                    \hline
     Mathematical / theoretical claims & Theoretical guarantees such as error bounds; analyses of computational complexity \\
     \hline
                            Dataset(s) & Any dataset(s) that are generated or used for any purpose, such as training or evaluating ML models \\
                            \hline
Empirical evaluation setup and metrics & Experimental setup including approaches to be compared, metrics, parameter settings \\
\bottomrule
\end{tabular}
\end{table*}

\begin{table*}[ht]
  \caption{The probing questions to answer included in the V1 prototype of REAL ML. Via our iterative design process, this table evolved into the activities included in the section on exploring limitations, as discussed in Section~\ref{sec:explore}.}
  \label{table:prototype-steps}
\begin{tabular}{p{5cm}p{9cm}}
\toprule
                \textbf{Step of Limitation Writing} &                                         \textbf{Prompt (answer these questions to write your limitation)} \\
\midrule
        Type of limitation (from areas list) & What type of limitation is this from the areas of limitations? Can you explicitly state the source of this limitation? \\
        \hline
              Broader impact of limitation & What is the broader (possibly negative if applicable) impact of this limitation on (1) your work? (2) the ML research community? (3) society? \\
              \hline
                    Alternatives & What are potential alternatives you could have taken instead of the choice that led to this limitation? \\
                    \hline
     How limitation was minimized & How did you minimize the impact of this limitation in your research? How do you plan to if you haven't yet? \\

\bottomrule
\end{tabular}
\end{table*}

\clearpage

\section{Stage 1 Interview Protocol}
As described in Section~\ref{sec:methods}, stage 1 of our study involved conducting semi-structured interviews with 20 ML researchers. The following interview protocol is the final version that was used to conduct these interviews. Small changes were made throughout stage 1 to ensure that the interview protocol matched the latest version of REAL ML. Specifically, Section \#2 of the interview protocol originally prompted participants to begin with types of limitations before considering sources of limitations, but later evolved to begin by asking participants to reflect on possible sources of limitations of their research before considering types of limitations, as described in Section~\ref{sec:tool}. When using the interview protocol, the interviewer personalized the follow-up questions and guidance on using REAL ML for each participant.

\subsection*{Section \#1}
\textbf{Goal: Get background information about the participant and their views of limitations}

\begin{enumerate}
    \item What areas of machine learning do you work on? Do you typically write theoretical or applied ML papers?
    \item Have you ever written a limitations section in a paper before? Why or why not? What was the experience like?
    \item What challenges (if any) have you faced before when it comes to identifying your assumptions and/or limitations in a research paper? When it comes to writing these things?
    \item How do you define a limitation? What do you think indicates a limitation has occurred in an ML research paper?

\end{enumerate}

\subsection*{Section \#2}

\textbf{Goal: Understand how ML researchers might use the tool, and ways that it can be improved}

\begin{enumerate}

\item[]\emph{Interviewer indicates that it is time to start reviewing the paper that the interviewee has brought to the interview. They explain that the next set of questions relates specifically to the research conducted in that paper. They mention ``Our intention is to reflect on the research process and paper writing process. There is no judgement here; this is a safe space.'' Begin with the first high level question to see what they say without showing them the tool.}

\item What kinds of assumptions did you make in your paper? What kinds of limitations did this research have?

\item[]\emph{Interviewer now shows the taxonomy ``questionnaire'' to the participant as a reference and lets them use it as a guide to refer to when thinking through the kinds of limitations of their ML research (making it very clear that this is a preliminary mockup and a very early version of the tool, and by no means will be the end tool). The participant selects one of their own research papers or a research paper that they are familiar with. They will be guided to the ``limitations'' of a ML research paper. The interviewer asks the following questions for each limitation type or for each decision type (depending on the interview). If each type is taking long, have the participant choose two or three types to focus on.}

    \item What is coming up for you in this limitation/decision category? What possible limitations can you think of that you might have encountered in this work?

\emph{If starting with limitations and participant is struggling to identify limitations, guide them to the ``Decisions'' sheet and prompt them there to see if that helps them remember the stages of research. If starting with decisions and participant is struggling to identify decisions, guide them to the ``Limitations'' sheet and prompt them there to see if that helps them remember what they may have encountered. When a new limitation from participant is encountered, interviewer guides participants to go through the ``Writing a Limitation'' prompts:}
    \begin{enumerate}
        \item What was the source of this limitation (e.g., an explicit decision, unforeseen consequences, lack of resources)?
        \item What were alternatives? Tradeoffs?
        \item What are the potential broader impacts of this limitation?
        \item What (if anything) did you do to mitigate this limitation? What could you have done?

    \end{enumerate}
    \item Was this limitation included or excluded from the original paper?
    \item[]\emph{If limitation was not included in participants' original paper:}
    \begin{enumerate}
        \item Were you aware of this limitation before publishing your paper? If not, if you had been made aware of this limitation while writing the paper, would you have included it?
        \item Would you feel comfortable including this limitation in your paper? Why or why not?
        \item Would this limitation be useful to state in your paper? Why or why not?
        \item Were there other reasons you chose to not include this in the original paper (e.g., practical reasons like page limits)?

    \end{enumerate}

\end{enumerate}

\subsection*{Section \#3}

\textbf{Goal: Improve the tool and make it more useful for ML researchers }
\begin{enumerate}
    \item{Improving the taxonomy:}
        \begin{enumerate}
            \item Which limitation types are missing / need improvement from this tool?
            \item Which decision-making types are missing / need improvement from this tool?
            \item How can the ``steps to writing a limitation'' be improved? What worked or didn't work for you? What was most/least helpful?
        \end{enumerate}

    \item When would it be helpful for you to use a tool like this (a more finalized and robust version) for your ML research (e.g., before research, during research, during writing, before submission)?
    \item In general, when do you think limitations should be stated in an ML research paper (e.g., beginning, middle, end)?
    \item Do you think that limitations are better expressed in one section? Or better disseminated throughout the entirety of a paper (e.g., when the limitation arose as a new idea is introduced in the paper)?
    \item In what ways did this tool confuse you or seem counterintuitive? \item In what ways could this tool be modified to be more useful to you as you conduct your own ML research?
\end{enumerate}

\section{Stage 2 Interview Protocol}

As described in Section~\ref{sec:methods}, stage 2 of our study
involved conducting interviews with six researchers who were
knowledgeable about ML and also experts in more sociotechnical
fields. Our goal in including these participants was to surface any
community norms or assumptions that might have been taken for granted
by ML researchers, but would stand out to ML-adjacent
researchers. These participants were asked to provide feedback on V2
or V3 of REAL ML.

\subsection*{Section \#1}
\textbf{Goal: Get background information about participant and their views on limitations}

\begin{enumerate}
    \item What areas of research do you typically focus on? What is your relation to the field of ML? What types of conferences do you publish in?
    \item What has your experience been towards limitations in your discipline(s)?
    \item What is your interpretation of ``limitations'' as they apply to ML research?
    \begin{enumerate}
        \item What do you think are some of the impact(s) of not reporting limitations in ML papers?
    \end{enumerate}
    \item How do you define the word limitation?
\end{enumerate}

\subsection*{Section \#2}
\textbf{Goal: Run through the tool and get feedback from participant}

\begin{enumerate}
    \item Feedback on limitations table
    \item Feedback on sources
    \item Feedback on tips/refining:
    \begin{enumerate}
        \item In the event of page limitations: Which limitations should be the focus? Which aspects of limitations should be the focus? (Feedback on the 4 qualities)
        \item When should limitations be in a paper?
        \item How should limitations be in a paper?
    \end{enumerate}
    \item Feedback on the table deliverable (other deliverables / output that would be better?)
    \item General thoughts / feedback on this tool?
\end{enumerate}

\section{Stage 3 Survey Protocol}

As described in Section~\ref{sec:methods}, stage 3 of our study
involved conducting an online survey with four ML
researchers. We sent a copy of REAL ML to each participant
via email and asked them to use it on their own, without any
additional guidance. After using REAL ML, each participant provided
feedback via an online survey, included below.\looseness=-1

\begin{enumerate}
    \item Please paste your final limitations section that you wrote at the end of the tool for your current ML research paper in the space below.
    \item Please rate your level of agreement or disagreement with the following statements about the tool (Strongly disagree, Disagree, Neither agree nor disagree, Agree, Strongly agree):
    \begin{itemize}
        \item The tool fostered a greater appreciation of the value of identifying, disclosing, and discussing the limitations of my research.
        \item The tool helped me identify limitations that I would not have identified otherwise.
        \item I felt better prepared to write about these limitations than I would have been without the tool.
        \item I was more willing to disclose and discuss the limitations in my paper than I would have been without the tool.
        \item I would use the tool again when writing research limitations of my work.
    \end{itemize}
    \item Did you choose not to disclose any identified limitations in your writeup? If so, can you tell us about them and explain why you chose not to disclose them?
    \item Which audience(s) did you have in mind when you prepared your limitations section with this tool?
    \item Did the tool help you uncover any limitations you weren't previously aware of? If so, please explain.
    \item Which activities and/or sections did you find the most helpful and why?
    \item Which activities and/or sections did you find the least helpful and why?
    \item If you have any other comments or suggestions about the tool, please write them here.
\end{enumerate}

\end{document}